\documentclass[conference, onecolumn]{IEEEtran} 

\IEEEoverridecommandlockouts
\usepackage{cite}
\usepackage{amsmath,amssymb,amsfonts}
\usepackage{algorithmic}
\usepackage{graphicx}
\usepackage{textcomp}
\usepackage{xcolor}
\usepackage{tikz}
\usetikzlibrary{graphs}
\usepackage{booktabs}
\usepackage{csquotes}

\usepackage{subfig}

\renewcommand{\O}[1]{\ensuremath{\mathsf{#1}}} 

\usepackage[ruled,vlined]{algorithm2e}
\SetKwInOut{Input}{Input}
\SetKwInOut{Output}{Output}
\SetKwProg{Fn}{Function}{}{}

\def\BibTeX{{\rm B\kern-.05em{\sc i\kern-.025em b}\kern-.08em
    T\kern-.1667em\lower.7ex\hbox{E}\kern-.125emX}}

\usepackage[margin=1.25in]{geometry}
    
\begin{document}

\newcommand\independent{\protect\mathpalette{\protect\independenT}{\perp}}
\def\independenT#1#2{\mathrel{\rlap{$#1#2$}\mkern2mu{#1#2}}}

\newtheorem{definition}{Definition}[section]


\title{
Optimal Transport for Fairness: Archival Data Repair using Small Research Data Sets\thanks{This work has received funding from the European Union’s Horizon Europe research and innovation programme under grant agreement No. 101070568. This work was also supported by Innovate UK under the Horizon Europe Guarantee; UKRI Reference Number: 10040569 (Human-Compatible Artificial Intelligence with Guarantees (AutoFair)).
}}

\author{

\IEEEauthorblockN{Abigail Langbridge\textsuperscript{\footnotesize{1}}\textsuperscript{\footnotesize{*}}}
\and
\IEEEauthorblockN{Anthony Quinn\textsuperscript{\footnotesize{1,2}}}
\and
\IEEEauthorblockN{Robert Shorten\textsuperscript{\footnotesize{1}}}

}

\maketitle

\renewcommand*{\thefootnote}{\fnsymbol{footnote}}
\footnotetext[1]{Corresponding author: al4518@ic.ac.uk}

\renewcommand*{\thefootnote}{\arabic{footnote}}
\setcounter{footnote}{0}
\footnotetext[1]{Dyson School of Design Engineering, Imperial College London}
\footnotetext[2]{Department of Electronic and Electrical Engineering, Trinity College Dublin}

\begin{abstract}
With the advent of the AI Act and other regulations, there is now an urgent need for algorithms that repair unfairness in training data. In this paper, we define fairness in terms of conditional independence between protected attributes ($S$) and features ($X$), given unprotected attributes ($U$). We address the important setting in which torrents of archival data need to be repaired, using only a small proportion of these data, which are $S|U$-labelled (the  research data). We use the latter to design optimal transport (OT)-based repair plans on interpolated supports. This  allows {\em off-sample}, labelled,  archival data  to be repaired, subject to stationarity assumptions. It also significantly reduces the size of the  supports of the OT plans, with correspondingly large savings in the cost of their design and of their {\em sequential\/} application to the off-sample data. We provide detailed experimental results with simulated and benchmark real data (the Adult data set). Our performance figures demonstrate effective repair---in the sense of quenching conditional dependence---of large quantities of  off-sample, labelled (archival) data.

\end{abstract}


\section{Introduction}
The notion of fairness is important across many applications, where some protected attribute should not impact some decision outcome. This concept of decision fairness is not new, with cases such as \cite{GriggsvDuke} and \cite{RiccivDeStefano} demonstrating strong legal precedence for avoiding discrimination from human decisions, even when protected characteristics such as race, sex or sexuality are not directly considered. For instance, in \cite{GriggsvDuke} the race of candidates was not directly considered, but could be inferred through high-school attendance and performance on tests standardised to median high-school graduates. Since black candidates were significantly less likely to hold a high-school diploma, these employment tests resulted in white candidates being almost ten times as likely to be considered for promotion \cite{blumrosen1972strangers}.

With decisions increasingly made by automated systems across a broad range of industries, from creditworthiness to automotive hazard detection, ensuring the fairness of these systems is crucial. These automated systems generally learn their behaviour from historic human decisions or from rules designed by domain experts. However, this can lead to systems encoding historical biases, or introducing bias through the learning process \cite{feldman2015certifying}.

Notably, the AI Act \cite{AI_act} marks a significant milestone in this regard, proposing a comprehensive framework to regulate AI systems and address potential risks, including bias and discrimination. One key stipulation of the AI Act is the requirement for transparency even for limited-risk applications. This has strengthened research effort towards AI fairness, including a significant body of work addressing the improvement of fairness through the \textit{repair} of data or models \cite{caton2020fairness, zafar2017fairness, besse2018confidence, friedler2019comparative}.

Many of these methods for data repair, however, rely on the notion that data are finite and static such that a repair operation can be designed and conducted once, and the problem of fairness is solved. In many other cases, protected attributes are not recorded, leaving significant volumes of historic data unusable with these SOTA repair schemes \cite{feldman2015certifying, gordaliza2019obtaining}, etc.). Evaluation is often conducted using benchmark data sets such as Adult Income \cite{adult} or COMPAS \cite{compas}, which by design are labelled, static and finite. However, these benchmark data sets do not represent real applications where data are harvested dynamically or sequentially. Calculating a new repair with every update to the data is likely to be prohibitively expensive \cite{friedler2019comparative, dvurechensky2018computational}. As such, we propose a repair operation inspired by approaches in domain adaptation \cite{courty2017joint} which can be learned on some small, labelled data set collected specifically for use in fairness repair, and then applied to repair historic, dynamic data without requiring that the repair be updated. 

The layout of the paper is as follows: in Section \ref{sec:modelling}, we define the notion of fairness in this setting and introduce the metrics used to evaluate it; in Section \ref{sec:related-work} we introduce related work in data repair with specific focus on \cite{gordaliza2019obtaining}; in Section \ref{sec:off-sample} we propose our method for off-sample data repair; in Section \ref{sec:experiments} we evaluate our proposed method on simulated and benchmark data; and in Sections \ref{sec:discussion} and \ref{sec:conclusion} we introduce avenues for future work and present the key conclusions of our work.

Throughout the paper, we denote random variables (r.v.s) by capital letters, $X$, and realizations of these random variables by lowercase letters, $x$, and often we do not need to distinguish notationally  between a r.v.\ and its realizations. We denote sets by $\mathbb{R}$, $\mathbb S$, etc., and functions by $\O{D}$, $\O{W}$, etc. We distinguish between observed and repaired data with  $x$ and $x'$, respectively. We make the usual assumptions about the probability space. The  probability model of a random variable, $X=x\in \mathbb{X}$, is denoted by $\O{F}(x)$ in the general case, being an element of the set of probability distributions with support, $\mathbb{X}$, denoted by ${\mathbb P}({\mathbb X})$. In the continuous case, $\O{F}(\cdot)$ is specified by its  probability density function (pdf), $f(x)$. In the discrete case, the probability mass function (pmf) of the probability model  is simply denoted by $\Pr[x]$.

\section{Fairness as Conditional Independence}\label{sec:modelling}
\begin{figure}
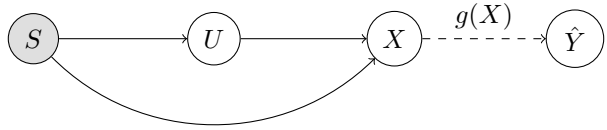

    \centering
    \tikz [scale=1.2] {
        \node (s) [circle, fill=gray!25, draw] at (0,0) {$S$};
        \node (u) [circle, fill=white, draw] at (2,0) {$U$};
        \node (x) [circle, fill=white, draw] at (4,0) {$X$};
        \node (y) [circle, fill=white, draw] at (6,0) {$\hat{Y}$};
        \graph { (s) -> (u) -> (x) ->[dashed, edge label=$g(X)$] (y), (s) ->[bend right=45] (x) };
        }
    \caption{Graphical representation of unfair data under the proposed $S$, $U$, $X$, $\hat{Y}$ model. Nodes in grey are unobserved, or may be unobserved.}
    \label{fig:graphical-model}
\end{figure}

The sample space, $\Omega$, of an  uncertain observational experiment, comprises elementary outcomes, $\omega \in \Omega$.  We define the following  r.v.s.\ in the usual way:
\begin{equation}
    Z = \{ X, S, U\}.
\end{equation}
Here, $X(\omega) = x \in \mathbb{R}^d$ is a length-$d$ feature vector, $S(\omega) = s \in \{0,1\}$ is a binary sensitive attribute, and $U(\omega) = u \in \{1, 0\}$ is a binary attribute which is not sensitive, and therefore unprotected. We assume that $X$ and $U$ are observed, and $S$ may be unobserved, as shown in Figure \ref{fig:graphical-model}. Each observation is classified accordingly to a specified rule: $g(X) : \mathbb{R}^d \rightarrow \{0,1\}$. 

To put this model in context, we consider the Adult Income data set \cite{adult}. In this setting, $S$ represents the protected attribute of race (which may or may not be observed), $U$ indicates whether the individual is educated to college level or above (observed), and $X$ are the remaining features (observations). By training on a set of such observations, we can design a prediction rule (classifier) of  whether  the annual salary of a future adult is greater than \$50,000 (i.e.\ $\hat{Y} =1$), or not. 

The joint probability model, $\O{F}(\cdot)$, induced by these r.v.s, is
\begin{equation}
    \O{F}(x,s,u) \equiv f(x \vert s, u) \Pr[s \vert u] \Pr[u],
\label{eq:prob_model}
\end{equation}
where $f_{s,u} \equiv f(x \vert s, u)$ is a class-conditional observation model (pdf) for the features, $x\in{\mathbb R}^d$, and the remaining factors are Bernoulli pmfs. 

We distinguish between two distinct sets of observations, a {\em research data set\/} where the labels, $S_R$, are observed, so that a typical (composite) observation\footnote{For ease of notation, we suppress the index into the specific observation we are considering.} is $z_R = \{ x_R, s_R, u_R\}$, and an {\em archival data set\/} where $S$ may not be observed\footnote{This assumption is considered in more detail in Section~\ref{sec:off-sample}.}, and a typical observation is  $z_A = \{ x_A, u_A\}$. The archival data may be observed online (sequentially), or they may be drawn from a prior-observed data set. We assume that the number of research data  is much smaller than the number of archival data, i.e.\ $n_R \ll n_A$, which together form an independent, identically distributed (iid) sample from $\O{F}(x,s,u)$ (Equation~\ref{eq:prob_model}). 

As an example, consider a job application setting. $n \equiv n_R + n_A$ applicants for a job  will all have provided details of their career achievements ($X$) and maximal educational level ($U$). A small number, $n_R$, of these may have volunteered to fill in an HR survey at the time of applying, indicating their religious affiliation and gender identity ($S$).

\subsection{ Conditional Independence: a sufficient condition for (sub-group) fairness}
\begin{figure}
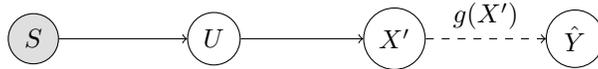

    \centering
    \tikz [scale=1.2] {
        \node (s) [circle, fill=gray!25, draw] at (0,0) {$S$};
        \node (u) [circle, fill=white, draw] at (2,0) {$U$};
        \node (x) [circle, fill=white, draw] at (4,0) {$X'$};
        \node (y) [circle, fill=white, draw] at (6,0) {$\hat{Y}$};
        \graph { (s) -> (u) -> (x) ->[dashed, edge label=$g(X')$] (y) };
        }
    \caption{Graphical representation of fair data under the proposed $S$, $U$, $X$, $\hat{Y}$ model, where the link between $S$ and the fairness-repaired data $X'$ is mediated by $U$.}
    \label{fig:graphical-model-fair}
\end{figure}

\noindent
\begin{definition}[$u$-conditional fairness] 
\label{def:fair}
We define $U$-conditional fairness as 
\[ (X \independent S) \vert U, \]
where $\independent$ denotes stochastic independence; i.e.\  $X$ is independent of the protected attribute, $S$, for each state, $u \in \mathbb{U}$, of the unprotected attribute (Figure \ref{fig:graphical-model-fair}).\hfill $\square$
\label{def:fairness}
\end{definition}
Fairness definitions in the literature emphasize {\em unconditional\/} independence between  outputs, $\hat{Y}$, and protected attributes, $S$; i.e.\ $\hat{Y} \independent S$ \cite{friedler2019comparative}.  Note that our {\em stochastic\/} definition, above, is a sufficient condition for classifier outcome fairness, since $\hat{Y} \equiv g(X)$ (Figure~\ref{fig:graphical-model-fair}). It is also sufficient for achieving fairness under various proxies, such as statistical parity and disparate impact~\cite{Hardt:23}. Other classical functional decoupling metrics~\cite{Gret:05}---based on cross-covariance operators---are also either necessary or equivalent (depending on the extent of the proposed decoupling) to Definition~\ref{def:fair}.    
Meanwhile, we opt for this {\em conditional\/} definition of fairness---a necessary condition for unconditional independence---in order to distinguish {\em structural (or societal) unfairness\/}, where $S \not{\!\! \independent} \; U$, from model (or AI) unfairness, where $(X \not{\!\! \independent} \; S) \vert U$. Our purpose is to identify and repair the latter only. Taking the  Adult Income data set~\cite{adult}, the fact that more white individuals are well educated than non-white individuals (structural/societal unfairness) is not something (regrettably) that is our business to repair. However,  the fact that white, well-educated individuals are predicted to earn more than non-white, well-educated individuals (model/AI unfairness) {\em is\/} something that we are concerned with repairing. This distinction is a key motivator for our definition.
Defining fairness in this way is also likely to uncover previously hidden under-representation bias. For instance, $s=0$ and $s=1$ may be equally probable in the marginal population but may not be equally distributed across groups $u \in \mathbb{U}$, meaning that  the sizes of the $s|u$-subgroups may vary with $u$.

\subsection{$u$-Conditional Fairness Metrics }\label{sec:fair-metrics}
Widely adopted fairness metrics (i.e.\ proxies) \cite{narayanan21fairness, Hardt:23} can all be redefined in conditionally independent terms, quantifying, for instance, the degree of subgroup fairness for $s$-indexed subgroups in each $u$-indexed group. 
Let us consider disparate treatment, perhaps  the most fundamental of the fairness metrics for classifier outputs \cite{zafar2017fairness}, which encodes the notion that the output should not depend on the sensitive attribute. Its $u$-conditional definition---necessary for our stochastic definition of fairness (Definition~\ref{def:fair})---is as follows:
\begin{definition}[Disparate Treatment]
    \[ \Pr(\hat{Y} = y \vert s, u) = \Pr(\hat{Y} = y \vert u), \: \forall (y,s,u) \in \{0, 1\}^3.\]
    \label{def:disparate-treatment}
\hfill $\square$
\end{definition}
Disparate impact (DI) is often adopted as the proxy for quanifying the extent to which Definition~\ref{def:disparate-treatment} is met. DI is one of the most widely used metrics due to its close relationship to legal literature in the United States and advocacy by the US Equal Employment Opportunity Commission \cite{useeoc_di}. The DI of a classifier, $\hat{Y}\equiv g(X)$ (Figure~\ref{fig:graphical-model}), in our $u$-conditional setting is defined as follows:
\begin{definition}[Disparate Impact]
    \[ \text{DI}(g,  u) \equiv \frac{\Pr(g(x) = 1 \vert S = 0, U = u)}{\Pr(g(x) = 1 \vert S = 1, U = u)}.  \]
    \label{def:disparate-impact}
\hfill $\square$ 
\end{definition}
If $\text{DI}(g,  u) = 1$, $\forall(g,u)$, then the classifier is unbiased. Meanwhile, if DI $> 0.8$, the classifier is considered to be fair \cite{useeoc_di}.

As with all functions of empirical probabilities, these proxies are subject to small-sample estimation errors \cite{besse2018confidence}, and are non-robust to the choice of the train-test split \cite{friedler2019comparative}. In this paper, we instead adopt a fairness measure for $X$---based on Kullback-Leibler divergence (KLD), $\O{D}[\cdot \vert \cdot ]$ \cite{KLD51}---that is agnostic to the decision rule, $g(\cdot )$ (Figure~\ref{fig:graphical-model}), and  is a function of the complete distribution (Equation~\ref{eq:prob_model}). 


First,  the $S$-dependence of the $u$-conditional distributions is quantified using the symmetrized Kullback–Leibler Divergence (KLD): 
\begin{definition}[$s|u$-dependence metric]
    \begin{align*}
    E_u \equiv &\frac{1}{2} \O{D} \biggl[ f(x | 0, u) \; || \; f(x | 1, u) \biggr] 
    + \frac{1}{2} \O{D} \biggl[ f(x | 1, u) \; || \; f(x | 0, u) \biggr]
    \end{align*}
\label{def:kld_fairness}
\end{definition}
Next, the $u$-expectation of $E_{u}$ is evaluated, yielding a fairness summary for the marginal model (being a mixture model) of $X$. 
A lower $E \geq 0$ represents fairer data:
\begin{equation}
    E = \sum_u \Pr[u] \; E_{u}
\label{eq:hat-kld-u}
\end{equation}
\hfill $\square$


\section{Optimal Transport for Data Repair}\label{sec:related-work}
There are two main approaches to achieving fairness in classification problems: either the classifier can be constrained during training, or the data can be modified, i.e.\ repaired \cite{zafar2017fairness, caton2020fairness}. Both methods lead to a reduction in classification performance, either by constraining the optimization search space or by reducing the predictive resource (e.g.\ correlation) available for prediction of the output, $\hat{Y}$, via the inputs, $X$ (Figure~\ref{fig:graphical-model}). In this work, we focus on data repair for fairness, in that we modify the data, $X$,  to $X'$, in a way that reduces its dependence on $S$, given $U$ (Figure~\ref{fig:graphical-model-fair}). 

Since we are seeking a repair operator that establishes conditional independence  between $X$ and $S$ (Definition~\ref{def:fairness}), it follows that our repair target should be $S$-invariant. {\em In tandem}, we want  this repair to the  minimal, i.e.\ the target should be close,  in some way, to both $s$-conditional distributions simultaneously ($\forall u$).  An optimal transport (OT)-based repair provides a compelling setting for this problem, and there are many works which consider OT for data repair \cite{dwork2012fairness, feldman2015certifying, gordaliza2019obtaining, chzhen2020fair}.

In the classical unregularized Kantorovich OT paradigm~\cite{peyr:19,Gali:16}, we specify the marginal probability measures (distributions), $\mu$ (on domain $\mathbb X$), and $\nu$ (on domain $\mathbb Y$). We  then design the unique joint distribution (i.e.\ a coupling, which we call the OT plan), $\pi^*$ (on the product domain $\mathbb{X}\times\mathbb{Y}$), which (i) has $\mu$ and $\nu$ as its marginals, and which (ii) minimizes the expected cost of transport (i.e.\ coupling) between $X$ and $Y$ under these marginal constraints. The  cost metric, $\mathsf{C}$,  must therefore also be specified {\em a priori}, for all transport paths, $(x,y)\in \mathbb{X}\times\mathbb{Y}$. Typically, $\mathsf{C}(x,y)\equiv ||x-y||_p^p$, $p\in\mathbb{N}^+$, i.e.\ $\mathsf{C}^\frac{1}{p}$ is the $\mathsf{L}_p$-norm on $\mathbb{X}\times\mathbb{Y}$. A key property of $\pi^*$ is that it induces the Wasserstein-$p$ metric in the space, $\mathbb{P}$, of probability measures, $\mu$ and $\nu$~\cite{peyr:19}. 

Noting, therefore, that the stochastic knowledge constraints (i.e.\ inputs) for OT are $\mu$ (called the {\em source}) and $\nu$ (the {\em target}),  it remains to associate these with appropriate conditional distributions in our mixture model (Equation~\ref{eq:mix}, Figure~\ref{fig:graphical-model}). $\mu$ is defined, in turn, as each {\em unrepaired\/} $s|u$-conditional component identified via the research data. $\nu$ is an appropriately defined $s|u$-independent {\em repaired\/} distribution, which is close---in the sense defined in Section~\ref{sec:bary}---to both source distributions. Our approach will follow~\cite{gordaliza2019obtaining}, but now adopting the $s|u$-conditional definition of fairness advocated in the current paper (Definition~\ref{def:fairness}).

 In what follows, we adopt a minimally opaque notation, via the following agreements:
 \begin{itemize}
 \item[(i)] $\mu$, $\nu$ and $\pi$ denote probability measures (distributions), without distinction as to their type. When their supports are continuous, we assume that the dominating measure is Lebesgue, in which case  $\mu$, $\nu$ and $\pi$ also denote the induced densities (i.e.\ Radon-Nikodym derivatives~\cite{Rao:18}). In the discrete case, involving counting measure,  $\mu$, $\nu$ and $\pi$  also denote the  probability mass functions (pmfs) of the respective measures.   \item[(ii)] We will also denote the empirical measures---specifically their probability mass functions---induced by random samples from the respective distributions by the same symbols, since the context makes clear when these empirical measures are in play. 
 \item[(iii)] Recalling that separate $u$-conditional repairs will be designed $\forall u\in \mathbb{U}$, wxe will suppress this $u$-conditioning in the notation, for the time being. 
 \end{itemize}

\subsection{Barycentric Repair Target}
\label{sec:bary}
In this section, we adopt the method proposed in \cite{gordaliza2019obtaining}, and apply it to  our $S, U, X, Y$ setting (Figure~\ref{fig:graphical-model}). 

Consider empirical distributions, 
$\mu_0, \mu_1 \in \mathbb{P}(\mathbb{R})$:  
\begin{equation}
\mu_s \equiv \frac{1}{n_s} \sum_i \delta_{x_{s,i}}, \;\;\;s\in{0,1}.\label{eq:emp} 
\end{equation}
Here, $x_{s,i}$ are the $s|u$-conditional observations from the research data set, $\mathbb{X}_R$, and $n_R \equiv n_0 + n_1$.


The optimal transport plan, $\pi^*$, transports points, $x_{0,\cdot} \in \mathbb{X}_0$,  from the $S=0$ class, to points, $x_{1,\cdot}\in \mathbb{X}_1$, in the $S=1$ class,  with minimal expected cost in respect of the specified cost function, $\mathsf{C}(x_0, x_1)$. In the Kantorovich formulation of OT\cite{kantorovich1942translocation}, this optimal repair plan is
\begin{equation}
    \pi^* \equiv \arg \min_{\pi \in \Pi(\mu_0, \mu_1)} \sum_{x_1} \sum_{x_0} \mathsf{C}(x_0,x_1) \pi(x_0, x_1),
    \label{eq:kantorovich_emp}
\end{equation}
where the couplings, $\pi$, are joint distributions (specifically, pmfs) over the product space, $\mathbb{X}_0 \times \mathbb{X}_1 $; i.e.  
\[
\Pi(\mu_0, \mu_1) \equiv \left\{ \pi \in \mathbb{P} (\mathbb{X}_0 \times \mathbb{X}_1) : \O{T}_{{x_0}_\sharp} \pi = \mu_0,\;\;  \O{T}_{{x_1}_\sharp} \pi = \mu_1 \right\}.
\]
Here, $\O{T}_{{x_0}_\sharp} \pi$ and $\O{T}_{{x_1}_\sharp}\pi$ are the push-forward measures induced by the operators,   $\O{T}_{{x_0}_\sharp}(x_0,x_1) \equiv x_0$ and $\O{T}_{{x_1}_\sharp}(x_0,x_1) \equiv x_1$, respectively.

The expected cost of transport in the optimal case (Equation~\ref{eq:kantorovich_emp}) defines a metric in the space, $\mathbb{P}(\mathbb{R})$ (where we assume $\mathbb{X}_0 \cup \mathbb{X}_1 \subseteq \mathbb{R}$). Specifically, if $\mathsf{C} \equiv \mathsf{L}_p^p$, $p\in\mathbb{N}^+$, then  
\begin{equation}
  \O{W}_p^p (\mu_0, \mu_1) \equiv \min_{\pi \in \Pi(\mu_0, \mu_1)} \sum_{x_1} \sum_{x_0} \mathsf{C}(x_0,x_1) \pi(x_0, x_1),
\label{eq:Wass}
\end{equation}
where $\O{W}_p$ is the $p$-Wasserstein distance. 

The Wasserstein barycentres fall on the geodesic, $\nu_t$, between the conditionals, $\mu_0$ and $\mu_1$, for $t \in [0,1]$ \cite{peyr:19}. These distributions represent the respective minimizers of the following $t$-indexed objective: 
\begin{equation}
\label{eq:baryt}
 \nu_t \equiv   \min_{\nu}  \left\{ (1 - t)\O{W}_p^p(\mu_0, \nu) + t\O{W}_p^p(\mu_1,\nu) \right\}.
\end{equation}
The $\O{W}_2$ barycentres (i.e.\ $p\equiv 2$)---which are of particular interest due to Brenier's theorem \cite{brenier1987decomposition}---arise in the case where the cost function is the squared Euclidean norm, i.e.\ $\O{C}\equiv \O{L}_2^2$. 
However, we note that---in our empirical setting---the conditions for Brenier's theorem are not met, since  $\mu_0, \mu_1$, are discrete measures on $\mathbb{X}_0$ and $\mathbb{X}_1$, respectively. 

To achieve a repair at equal expected cost---in the $p\equiv 2$ case---to both classes, $s \in \mathbb{S}\equiv \{1,0\}$, we are interested in the $t=0.5$ barycentre, i.e.\ $\nu_{0.5}$, which is the distribution where $\O{W}_2(\mu_0, \nu_{0.5}) = \O{W}_2(\mu_1,\nu_{0.5})$, i.e.\ the centre of the geodesic (Equation~\ref{eq:baryt}). In this case, we drop the the $0.5$ subscript, and denote this `fair barycentre'  simply as $\nu$. This repair target
is widely adopted as the fair target design  \cite{feldman2015certifying, chzhen2020fair}.  


\subsection{Geometric Repair}
\label{sec:geom}
Given the barycentric target, $\nu$, the following repair method is proposed in~\cite{gordaliza2019obtaining}, being  a generalization of a repair for one-dimensional data in \cite{feldman2015certifying}. 

In this setting, the barycentre need not be explicitly calculated. Instead, each point, $x_{s,\cdot}$, is transported to a new target point, $x'_{s,\cdot}$, in the support of the fair barycentre, $\nu$, according to the following mapping governed by the OT plan, $\pi^*$ (Equation~\ref{eq:kantorovich_emp})~\cite{gordaliza2019obtaining}:
\begin{equation}
    {x'}_{0,i} = (1-t) x_{0,i} +  n_0 t \sum_j \pi^*_{ij} x_{1,j}
\label{eq:gord_geom0}
\end{equation}
\begin{equation}
    {x'}_{1,j} = n_1 (1-t) \sum_i \pi^*_{ij} x_{0,i} + t x_{1,j}
\label{eq:gord_geom1}
\end{equation}
%
This method is unsuitable for off-sample repair since the transport is designed point-wise; i.e.\ only the on-sample points, $x_{0,i}$ and $x_{1,j}$, in the research data set, $\mathbb{X}_R$ (Section~\ref{sec:modelling}), are in the domains of the two repair operators above. Hence, they  cannot be used to repair off-sample points in the archival data set, $\mathbb{X}_A$. 

\section{Off-Sample Data Repair} \label{sec:off-sample}
We now propose a framework for $s|u$-indexed fairness correction of {\em archival data}, $x_A \in \mathbb{X}_A$, using the $s|u$-labelled research data, $x_R \in \mathbb{X}_R$ (Section~\ref{sec:modelling}). For the reason given in the  previous paragraph, we also refer to these as off-sample and on-sample data, respectively. $\mathbb{X}_R$ can be considered analogous to the `training set', i.e. we design our OT-based repair using these data. If then used to repair $\mathbb{X}_R$, we call this  \textit{on-sample repair}. In this sense, $\mathbb{X}_A$ can be considered to be the `test set', i.e.\ $\mathbb{X}_A$ are not available when designing the repair, but are then repaired by it, a process we call  \textit{off-sample repair}.

Recall that the aim of our repair is to achieve conditional independence between $x'$ (and therefore $\hat{y}'\equiv g(x')$ (Figure~\ref{fig:graphical-model-fair})) and $s$ for each $u \in \mathbb{U}$, following our conditional independence definition of fairness (Definition \ref{def:fairness}). We impose the following requirements:
\begin{enumerate}
    \item  As already stated in Section~\ref{sec:related-work}, the repair plan, $x\rightarrow x'$, should be minimally damaging with respect to the prediction of (off-sample, unrepaired) $x \sim \O{F}(x|s,u)$ (Equation~\ref{eq:prob_model}). For this reason, we adopt the barycentric repair with $t=\frac{1}{2}$ (Equation~\ref{eq:baryt}). 
    \item The repair plan designed on the research data set,  $\mathbb{X}_R$, should generalize to the repair of off-sample points, $x_A\in\mathbb{X}_A$, drawn from the same (stationary) population.
    \item The method should be computationally efficient, so that large data sets can be repaired.
    \item The repair scheme should be $u$-indexed, so that it is re-designed for each $u$-indexed group, $\mathbb{X}_u$. 
    \item $x_R \in \mathbb{X}_R$ are $s$-labelled by definition. We also assume that $x_A \in \mathbb{X}_A$ are $s$-labelled.
    \end{enumerate}

Regarding this final requirement, it is usual that the sensitive attributes, $S=s$, of archival points, $x_A$, are {\em not\/} available (measured) {\em a priori}, necessitating the identification of each  $u$-conditional mixture model, 
\begin{equation}
\O{F}(x|u) \equiv \sum_s \O{F}(x|s,u)\Pr[s|u],\;\;\;\;\; \forall u\in\mathbb{U},
\label{eq:mix}
\end{equation}
via standard methods, and the associated estimation, $\hat{s}|u$, of the $s|u$ labels of $x_A \in \mathbb{X}_A$~\cite{BishopMLbook06}.  This  task is not the focus of the present work, and its implementation does not disturb the repair methods we are proposing here. 
For these reasons, we will not distinguish between $\hat{s}|u$ and $s|u$ in the sequel. Further comment on this point is provided in Section~\ref{sec:discussion}.
 

To fulfil the five requirements above, we now propose a \textit{distributional repair} which is suitable for  off-sample points, $x_A$. The repair is first designed using the research data set, $\mathbb{X}_R$ (Section \ref{sec:our-method}: Algorithm~\ref{alg:dist_repair}), which can be considered as an initialization step analogous to model training. Then, off-sample data in $\mathbb{X}_A$ can be repaired online (Section \ref{sec:repair-off-sample}: Algorithm~\ref{alg:off-sample}) with modest computational overhead.

\subsection{Distributional Repair}\label{sec:our-method}
In our current approach, we stratify the repair by feature, $x_k \in x$, $k\in\{1,\ldots, d\}$ (as well as by $u\in{\mathbb U}$, following  Definition \ref{def:fairness}). Since OT in high dimensions is computationally prohibitive \cite{shirdhonkar2008approximate, petrovich2022feature, fan2022complexity}, we propose  this feature split in order to improve the scalability of the method and to avoid the curse of dimensionality. Throughout this section, we suppress the $u,k$ indexing for notational convenience, since the same procedures (see Algorithms \ref{alg:dist_repair} and \ref{alg:off-sample}, below) are applied for all $u \in \mathbb{U}$ and $k \in \{1,\ldots, d\}$. The $(u,k)$-stratification is clearly indicated in the algorithms.

\subsubsection{Interpolation of the empirical pmfs, $\mu_s$} 
\label{sec:interp}
To facilitate repair of $s$-indexed off-sample points in the archive,  i.e.\ from the set $\mathbb{X}_{A,s} \backslash \mathbb{X}_{R,s}$, $\forall s\in\mathbb{S}$, we interpolate the empirical marginals, $\mu_0$ and $\mu_1$, uniformly across the range of (combined) $\mathbb{X}_R$, yielding the support set, $\mathbb{Q}$, for each marginal with $n_Q$ states. In Section \ref{sec:internQ}, we will investigate the influence of $n_Q$ on the performance of our repairs.
The pmfs of the interpolated marginals are computed via kernel density estimation,
\begin{equation}
    p_{s,q} \equiv \Pr[X=q | S=s]  \propto \sum_{i} {\mathsf K} (q - x_{R,s,i}, \;h),\;\;\;q\in\mathbb{Q},
\label{eq:kde_pmf}
\end{equation}
using the Gaussian kernel,
\begin{equation}
    {\mathsf K}(x, h) \propto \exp{\left(-\frac{x^2}{2h^2} \right)},
    \label{eq:KDEker}
\end{equation}
where the bandwidth parameter, $h$, is set using Silverman's method \cite{silverman1986density}.

In this work, we make the simplifying assumption that $n_{Q_{0}} \equiv n_{Q_{1}}$ such that the barycentre is represented on the same support as the two marginals and the quantization error on both $s$-conditional distributions is equal.

The complexity in calculating the repair using unregularised OT is now  $\mathcal{O}({n_Q}^3 \log{n_Q})$ \cite{rubner2000earth, dvurechensky2018computational}. For regularised methods, the  complexity of the Sinkhorn-Knopp algorithm \cite{sinkhorn1967concerning} for an  $\epsilon$-approximation of the OT plan is  $\mathcal{O}({n_Q}^2 / \epsilon^2)$ \cite{le2021robust, cuturi2013sinkhorn}. This removes the dependence on the potentially unbounded data cardinality, $n\equiv n_S + n_A$, and can  significantly reduce the complexity of the repair operation. The main active assumption is that $\mathbb{X}_R$ is a representative subset of the stationary composite dataset, $\mathbb{X} \equiv \mathbb{X}_A + \mathbb{X}_R$. We will explore these issues in detail in Section~\ref{sec:simulation}.

\subsubsection{Repair Design}
We define the repair target to be $\nu\equiv\nu_{0.5}$, i.e.\ as the $t=0.5$ barycentre along the $p=2$-Wasserstein geodesic (Equation~\ref{eq:baryt}) between the {\em interpolated\/} marginals (defined via Equation~\ref{eq:emp}, and simply denoted by $\mu_0$ and $\mu_1$, respectively).
 $\nu$ has the same support, $\mathbb{Q}$, as these interpolated marginals,  and so the $s$-indexed OT couplings, $\pi_s$, between $\mu_s$ and $\nu$  are  joint distributions over the product space of the marginal supports; i.e.\  $\pi_s \in \mathbb{P}(\mathbb{Q} \times \mathbb{Q})$: 
 \begin{equation}
    \pi^*_s \equiv \arg \min_{\pi \in \Pi(\mu_s, \nu)} \sum_{q_j\in\mathbb{Q}} \sum_{q_i\in\mathbb{Q}} \mathsf{C}(q_i,q_j) \pi(q_i,q_j). 
    \label{eq:kantorovich_trans}
\end{equation}
  If $n_Q \rightarrow \infty$, continuous limits of $\mu_s$ and $\nu$ are attained. With squared Euclidean cost $\mathsf{C}(q, q)\equiv ||q_i -q_j||_2^2$, the sufficient conditions specified by Brenier's theorem are met, and so the optimal Kantorovich plan, $\pi^*_s$, converges to a Monge map \cite{brenier1987decomposition, peyr:19}.

Note, in Algorithm \ref{alg:dist_repair}, the differences in how we construct the interpolated marginals, $\mu_0$ and $\mu_1$,  and the $s$-indexed transport plan, $\pi^*_s$, compared to the geometric repair in \cite{gordaliza2019obtaining}.

\begin{algorithm}
\DontPrintSemicolon
$\forall (u,s,k)\in \mathbb{U}\times \mathbb{S} \times 
\{1,\ldots,d\}$

\Input{
$(u,s)$-labelled research data: $\mathbb{X}_{R,u,s} \equiv \{x_{R,u,s,i},\; i=1,\ldots,n_{R,u,s}\} \subset\mathbb{R}^d$  \\ 
Resolution parameters: $ n_{Q,u,k} \in \mathbb{N}$ 
}

\Output{
$(u,s,k)$-indexed OT-repair plans: $\pi^*_{u, s, k}$ \\
Interpolated marginal supports: $\mathbb{Q}_{u,k}$ 
}
\RestyleAlgo{ruled}

\Fn{Design Repair$(\forall (u,s,k): \mathbf{x}_{R,u,s},  n_{Q,u,k})$}{
\nl \ForEach{$u \in \mathbb{U}$}{
  \nl \ForEach{$k \in \{1,\ldots, d\}$}{   
    \nl \ForEach{$i \in \{1,\ldots, n_{Q,u,k}\}$}{ 
        \nl  $\zeta_i=\frac{n_{Q,u,k}-i}{n_{Q,u,k}-1}\cdot\min({\mathbf x}_{R,u,k})$ $+\frac{i-1}{n_{Q,u,k}-1}\cdot\max({\mathbf x}_{R,u,k})$}
        \nl $\mathbb{Q}_{u,k} = \{\zeta_1, \dots, \zeta_{n_{Q,u,k}}\}$ \\
        \nl $M_{u,k} = \mathsf{C}(\mathbb{Q}_{u,k}, \mathbb{Q}_{u,k}) \in \left(\mathbb{R}\cup\{0\}\right)^{n_{Q,u,k}\times n_{Q,u,k}}$ [cost matrix]\\
        \nl \ForEach{$s \in \mathbb{S}$}{
            \nl Compute $p_{u, s,k} $ using Equation \eqref{eq:kde_pmf} \\
        }
        
        \nl Compute $\nu_{u,k}$ using Equation \eqref{eq:baryt} \\

        \nl \ForEach{$s \in \mathbb{S}$}{
            \nl Compute $\pi^*_{u,s,k}$ using Equation \eqref{eq:kantorovich_trans} \\
        }
    }
}
}
\caption{On-Sample Design of Distributional Repair Plan }
\label{alg:dist_repair}
\end{algorithm}

\subsection{Off-Sample (i.e.\ Archival) Repair} \label{sec:repair-off-sample}
Given the research-data-based optimal plans, $\pi^*_s$, $s \in \mathbb{S}$, , and the interpolated support of each marginal, $\mathbb{Q}$, both computed via Algorithm~\ref{alg:dist_repair}, we must relate labelled archive samples, $x_{A,s}$, to the corresponding interpolated marginal pmf, $\mu_s$. Almost surely (a.s.) in the continuous feature case, $x_{A,s}\notin \mathbb{X}_{R,s}$, but---under the stationarity assumption---we  assume only that $x_{A,s}\stackrel{\mathsf{iid}}{\sim} \mu_s$ in the same way that the labelled research data, $x_{R,s}$, are. We also assume that $x_{A,s}$ is in the range of $\mathbb{X}_R$.   
We suppress the $A$-subscript in the sequel, without risk of confusion. Denoting the round-down state of $x_{s,i}$ in $\mathbb{Q}$ by $\zeta_{q_{s,i}} \equiv \lfloor x_{s,i} \rfloor$, $q_{s,i} \in \{1,\ldots, n_Q\}$, we evaluate the following ratio:
\begin{equation}
    \tau_{s,i} = \frac{x_{s,i} - \zeta_{q_{s,i}}}{ \zeta_{q_{s,i} +1} - \zeta_{q_{s,i}}} \in (0,1).
\label{eq:support_interp}
\end{equation}
This is used  to define the first source of randomness in our archival repair algorithm (Algorithm~\ref{alg:off-sample}): a Bernoulli trial, $(\mathcal B)(\cdot)$, defined by $A_{s,i} \sim \mathcal{B}(\tau_{s,i})$. The realization, $a_{s,i} \in \{0,1\}$, of this trial then specifies that---of the two neighbouring elements of $x_{s,i}$ in $\mathbb{Q}$---it is the $q_{s,i} + a_{s,i}$ one which will be used in the repair of $x_{s,i}$. This corresponds to choosing row $q_{s,i} \leftarrow q_{s,i} + a_{s,i}$ of $\pi^*_s$.

In order to maintain the cardinality of the labelled archive sets, $\mathbb{X}_{A,0}$ and $\mathbb{X}_{A,1}$, we implement the mass-split dictated by the  Kantorovich OT plan, $\pi^*_{s}$, via a random  draw from the $n_Q$-state conditional multinomial pmf, ${\mathcal M}(\cdot)$, specified by normalizing the row of the plan selected randomly, as explained in the previous paragraph. This constitutes a   second source of randomness in the archival repair procedure, and avoids the deterministic mass splitting proposed in \cite{gordaliza2019obtaining}. In summary, the $s$-labelled archive data are randomly repaired by optimally transporting them, $x_{s,i} \rightarrow x'_{s,i}$, with distribution

\begin{equation}
    \Pr \Big[ x'_{s,i} = \zeta_j \big\vert x_{s,i} \Big] \propto \pi^*_{q_{s,i},j}, \;\;1 \leq q_{s,i},j \leq n_Q.
    \label{eq:support_target}
\end{equation}
The resulting procedure for repairing the archive data is summarized in    Algorithm \ref{alg:off-sample}.
\begin{algorithm}
\DontPrintSemicolon
$\forall (u,s,k)\in \mathbb{U}\times \mathbb{S} \times 
\{1,\ldots,d\}$

\Input{$(u,s)$-labelled archive data: $\mathbb{X}_{A,u,s} \equiv \{x_{A,u,s,i},\; i=1,\ldots,n_{A,u,s}\} \subset\mathbb{R}^d$ \\ 
$\mathbb{Q}_{u,k}$\\ 
$\pi^*_{u,s,k}$
}

\Output{$s$-conditionally-independent, $u$-labelled, repaired archival data, $\mathbb{X}'_{A,u}$}
\RestyleAlgo{ruled}

\Fn{Off-Sample Repair$(\mathbb{X}_{A,u,s}, \; \mathbb{Q}_{u,k}, \;\pi^*_{u,s,k}) $}{
\nl \ForEach{$u \in \mathbb{U}$}{
$\mathbb{X}'_{A,u} = \emptyset$\\
\nl \ForEach{$s \in \mathbb{S}$}{
    \nl \ForEach{$k \in \{1,\ldots, d\}$}{
        \nl \ForEach{$i \in  \{1,\ldots,n_{A,u,s}\}$}{
            
            \nl $ \zeta_q = \lfloor x_{A,u,s,k,i}   \rfloor \in \mathbb{Q}_{u,k}$\\
            \nl Compute $\tau$ using Equation \eqref{eq:support_interp} \\
            \nl $a  \sim \mathcal{B}(\tau)$ \\
            \nl $q = q + a$ \\
            \nl Compute $x'_{u,s,k,i}$ by simulating via    Equation \eqref{eq:support_target} 
 }       
 }
   \nl   $\mathbb{X}'_{A,u} = \mathbb{X}'_{A,u} \cup  
   \{x'_{u,s,k,i}\}$
       }
    }
}
\caption{Off-Sample Data Repair. Note that it takes the outputs of Algorithm~\ref{alg:dist_repair} among  its inputs. }
\label{alg:off-sample}
\end{algorithm}


\section{Simulation and Real-Data Studies}\label{sec:experiments}
We now conduct a number of experiments to validate our approach to archival (off-sample) data repair via research-data-trained (on-sample) OT repair operators. In Section \ref{sec:simulation}, we design a simulation study which we use to validate the notion of off-sample repair, and we study the operating conditions, $n_R$  and $n_Q$ of Algorithm~\ref{alg:dist_repair}.
Then, in Section \ref{sec:adult}, we apply our method to the Adult income data set \cite{adult} to validate its performance on large, noisy data for which the $(u,s)$-conditional pmfs must be estimated from the on-sample data, and the $s|u$ labels must be estimated for the off-sample (archival) data. 

\subsection{Simulation Study}\label{sec:simulation}
We simulate our composite (i.e.\ research and archival) data set (with $d=2$) as follows:
\[
x_{u,s} \stackrel{\mathsf{iid}}{\sim} {\mathcal N}(\mu_{u,s}, \Sigma_{u,s}), \;\;\;i=1, \ldots, n\equiv n_R + n_A \equiv 5,500.
\label{eq:GMM}
\]
The means of the $(u,s)$-conditional normal components are 
$\mu_{0,0} \equiv [-1,-1]^T$,
$\mu_{0,1} \equiv [0,0]^T$,
$\mu_{1,0} \equiv [1,1]^T$, 
$\mu_{1,1} \equiv [0,0]^T$,  and  
 $\Sigma \equiv  I_2$ (the $2\times 2$ identity matrix). 
We enforce balance---i.e.\ we avoid under-representation bias \cite{mehrabi2021survey}---in the $u$-indexed populations via $\Pr(u = 0)\equiv 0.5$. The $s$-splits of these populations are then chosen to be $\Pr[s=0|u = 0]\equiv 0.3$ and $\Pr[s=0|u = 1]\equiv 0.1$, ensuring that  the  $s=1$ sub-population is the dominant one in each $u$-indexed population. Initially, we split the data set into $n_R \equiv 500$ research (on-sample) points, and $n_A \equiv 5000$ archival (off-sample) points. We set $n_{Q_{u,s}} \equiv 50 \; \forall \; u, s \in \{0,1\}^2$ for simplicity since the $u$-conditional distributions are translations of one another.

The $(u,s)$-labels of the research points are observed, so that $\mathbb{X}_R$ is partitioned into subsets, $\mathbb{X}_{R,u,s}$, as indicated at the input of 
Algorithm~\ref{alg:dist_repair}. Similarly, we assume that the archival data are also $(u,s)$-labelled, or, if not, these labels can be estimated accurately. For more comment on this archival classification issue, please see Section~\ref{sec:discussion}.   As explained in Section \ref{sec:off-sample}, we design our OT repairs for each $ (u,s)$ group, and for each feature (channel) in turn, $k\in\{1,\ldots, d\}$, using only the research data set, $\mathbb{X}_R$. We then use these to repair each channel, $k$, of the $(u,s)$-labelled archive data, $\mathbb{X}_A$. We evaluate the repair performance for both data sets, $\mathbb{X}_R$ and $\mathbb{X}_A$, via the $E$ metric (Definition \ref{def:kld_fairness}) over 200 independent Monte-Carlo simulations.

\subsubsection{Off-Sample Repair}
\label{sec:off-sam-sim}
We evaluate our data repair method on the research data set, $\mathbb{X}_R$, and archival data set, $\mathbb{X}_A$, separately. This allows us to evaluate the effectiveness  of off-sample vs.\ on-sample repairs, i.e.\ how well we can repair archival data (Algorithm~\ref{alg:off-sample}) which were not available at the design stage of the OT repairs (Algorithm~\ref{alg:dist_repair}), analogously to the assessment of generalization in a  machine learning model subject to a particular train-test split. Table \ref{tab:sim-repair} shows the $E$ performance measure for each feature dimension, $k$, marginally w.r.t.\ the $(u,s)$ groups, for $\mathbb{X}_R$ and $\mathbb{X}_A$, respectively. 

The performance of our {\em distributional\/} OT-repair method is captured by comparing the $E$ performance figures for the repaired data sets to those of the unrepaired data sets.  Furthermore, in the case of the research data, $\mathbb{X}_R$, we benchmark our performance against that of the 
{\em geometric\/}  OT-repair method in \cite{gordaliza2019obtaining}. Recall that the latter is a repair for on-sample data only (Section~\ref{sec:geom}), and so cannot be used in the repair of $\mathbb{X}_A$, as indicated in Table~\ref{tab:sim-repair}. 
While the geometric repair provides a slightly more $s$-invariant repair (lower $E$) compared to our distributional repair, both perform  similarly and effectively. We argue that the ability of the latter to repair (potential torrents of) off-sample data is of great importance in real data contexts.   

\begin{table}
    \centering
    \caption{OT-based repairs (quenching of conditional dependence) for simulated data (two bivariate Gaussian sub-groups). Comparison of our {\em distributional\/} OT-based repair method to the {\em geometric\/} OT-based repair in \cite{gordaliza2019obtaining}. Lower figures indicate improved repair.}
    \begin{tabular}{l|cc|cc}
        \toprule
        Repair & \multicolumn{2}{c|}{$E_{k}$ (Research)} & \multicolumn{2}{c}{$E_{k}$ (Archive)} \\
         & $k=1$ & $k=2$ & $k=1$ & $k=2$ \\
         \midrule
         None & $7.486 \pm 1.445$ & $7.271 \pm 1.475$ & $6.279 \pm 0.543$ & $6.377 \pm 0.522$\\ 
         Distributional (ours) & $0.0899 \pm 0.0373$ & $0.0926 \pm 0.0368$ & $\mathbf{0.3926 \pm 0.1671}$ & $\mathbf{0.4443 \pm 0.2028}$\\
         Geometric \cite{gordaliza2019obtaining} & $\mathbf{0.0071 \pm 0.0030}$ & $\mathbf{0.0073 \pm 0.0033}$ & - & -\\
         \bottomrule
    \end{tabular}
    \label{tab:sim-repair}
\end{table}

\subsubsection{Operating Conditions, $n_R$ and $n_Q$}\label{sec:opcon}
Two key considerations in the application of our proposed repair method are the chosen size of the research data per $(u,s)$-labelled group, i.e.\ $n_{R,u,s}$ (Algorithm~\ref{alg:dist_repair}), and the chosen   number of states, $n_{Q,u,k}$, in the uniformly interpolated support of each estimated  marginal pmf, $\mu_{u,s,k}$.  Recall that the same interpolation support, $\mathbb{Q}_{u,k}$, is adopted for each $s$-indexed subgroup, $\mathbb{X}_{R,u,s,k}$, and that a distinct interpolated OT-repair is designed for each feature/channel, $k\in\{1,\ldots, d\}$, in turn (Algorithm~\ref{alg:dist_repair}). 

\begin{figure}
    \centering
    \includegraphics[width=\columnwidth]{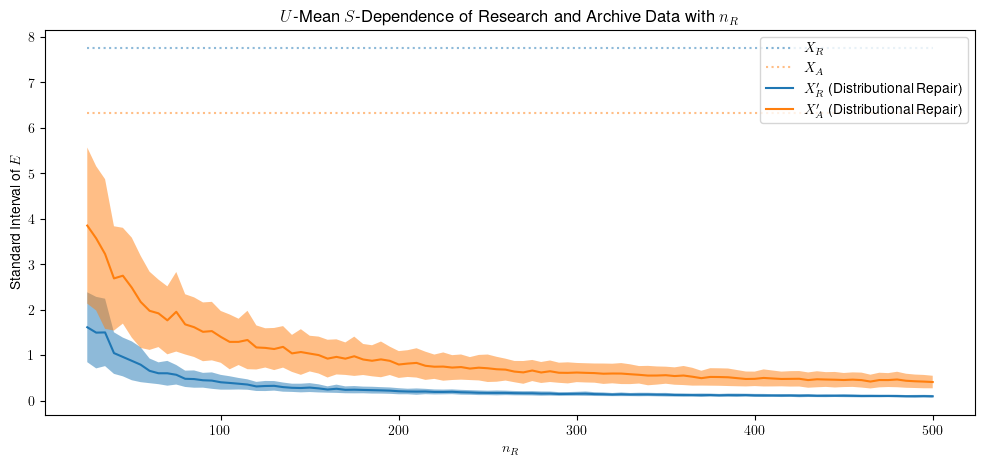}
    \caption{Simulated bivariate Gaussian sub-groups (Section~\ref{sec:simulation}).   Empirical approximation of $E$
    (Equation \ref{eq:hat-kld-u}) as the size of the research data set, $n_R$, increases. For this experiment, $n_A = 5000$ and $n_Q = 50$.}
    \label{fig:n_R}
\end{figure}

\paragraph{Research Data Size, $n_{R}$}
We consider the $E$ performance figure (Definition~\ref{def:kld_fairness}, aggregated over both features, $k\in\{1,2\}$) for the research and archival data as the net size of the research data, $n_R\equiv \sum_{u,s} n_{R,u,s}$, increases from 25  to $750$ observations. As Figure \ref{fig:n_R} shows,  $E$ converges to a steady-state value for both the research and archival data, even for research data set sizes as small as 10\% (i.e.\ $n_R\equiv 500$) of the archival data set size, $n_{A}\equiv \sum_{u,s} n_{A,u,s}$. 
This indicates that  the per-feature empirical $(u,s)$-conditional distributions, $\tilde{f}(x_k | \mathbf{x}_{R,u,s})$, have converged---in the sense of the laws of large numbers---to the   underlying per-feature distributions, $\mu_{u,s,k}$, for each $u\in\mathbb{U}$, $s\in \mathbb{S}$ and $k\in \{1,\ldots, d\}$. 

Note that the $E$ of the archival data remains higher than the research data due to the fact that they are off-sample with respect to the research data, ${\mathbf x}_R$, used to design (i.e.\ train) the OT repairs. Additionally, these archival observations may fall outside the range of the research data (and, therefore, of the interpolated supports, $\mathbb{Q}_{u,k}$). These potential artefacts are a consequence of the incomplete learning of the underlying marginals, $\mu_{u,s,k}$, via the research data. In real-data applications (e.g.\ in the next section), any statistical drift (i.e.\ nonstationarity) that may be present in the (torrents of) archive data will also be reflected in a suppressed repair performance, via $E$. These considerations point to the more stressful regime of repair represented by the off-sample vs.\ the on-sample case. Nevertheless, we still achieve a major reduction in conditional dependence on $s$, as seen by comparing with $E$ of the complete unrepaired data set (Figure~\ref{fig:n_R}). 

\paragraph{Resolution of the Interpolated Marginal pmfs, $n_Q$}
\label{sec:internQ}
We revert to the data sizes of the first simulation study (Section~\ref{sec:off-sam-sim}, i.e.\ $n_R\equiv 500$ and $n_A\equiv 5000$), a setting for which our repair method achieves excellent performance at $n_Q=50$ (Table~\ref{tab:sim-repair}). We now investigate the influence of the marginal interpolations on the performance of our repairs, via $n_Q$. For convenience, we assign these numbers equally across all $u$-labelled groups (as well for each $s\in\mathbb{S}$, as explained in Section~\ref{sec:interp}), and across all features (channels), $k$. Recall that the range of each pmf support is set independently of $n_Q$ (see line 4 of Algorithm~\ref{alg:dist_repair}), and so increasing $n_Q$ increases the resolution of estimation of the underlying marginals, $\mu_{u,s,k}$, proportionally. We are interested in quantifying---via $E$ (Definition~\ref{def:kld_fairness}) for the composite  data, ${\mathbb X}_R \cup {\mathbb X}_A$---the effect of $n_Q$  on the performance of our distributional OT-repair scheme.   

\begin{figure}
    \centering
    \includegraphics[width=\columnwidth]{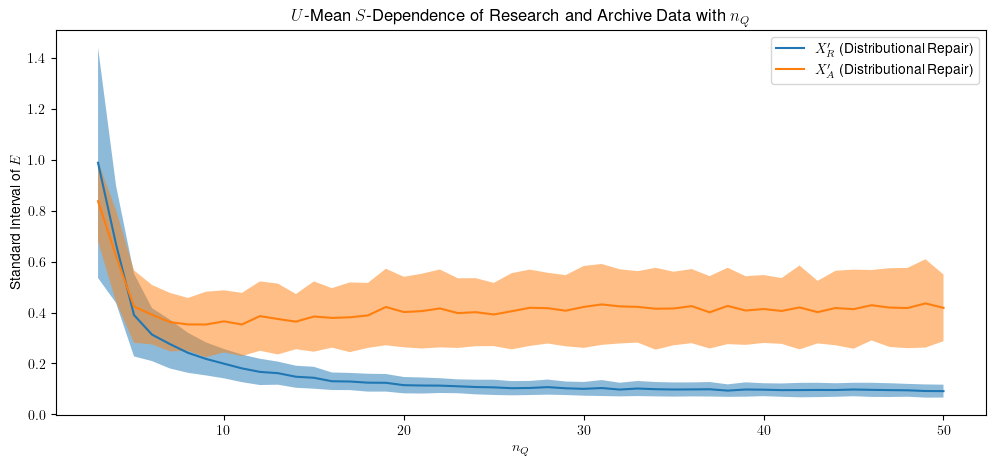}
    \caption{Simulated bivariate Gaussian sub-groups (Section V-A). Empirical approximation of $E$
    (Equation \ref{eq:hat-kld-u}) as $n_Q$ increases for the composite repaired data set. For this experiment, $n_R = 500$ and $n_A = 5000$.}
    \label{fig:n_Q}
\end{figure}

Results for $n_Q \in \{5, ..., 50\}$ are shown in Figure \ref{fig:n_Q}. Above a threshold, $n_Q\approx 30$, the repair performance converges. We conclude the following:
\begin{itemize}
    \item[(i)] Above this threshold, the statistics of the interpolated marginals, $p_{u,s,k}$ (Equation~\ref{eq:kde_pmf}), converge to those of  the underlying marginals, $\mu_{u,s,k}$, therefore also ensuring convergence of  the OT repair schemes, $\pi^*_{u,s,k}$ (Equation~\ref{eq:kantorovich_trans});
    \item[(ii)] At convergence, the number of pmf interpolants, $p_{u,s,k}$, is an order of magnitude fewer than the number of research (i.e.\ training) points, $n_R \equiv 500$. These interpolants function as pseudo-sufficient statistics for estimation of the marginals. It is interesting to note the significant compression which can be tolerated in our current setting, involving bivariate Gaussian marginals (Equation~\ref{eq:GMM}), which can be estimated well via KDE with Gaussian kernels (Equation~\ref{eq:KDEker}). This compression affords a major computation saving since the complexity of calculating unregularised OT plans relates to the cube of the support, while also generalizing for off-sample points, as we have seen;
    \item[(iii)] As stated above, the repair metric, $E$, is statistically invariant above $n_Q\approx 30$ for both $\mathbb{X}_R $ and $\mathbb{X}_A$. We anticipate that $n_Q$ will have more impact on \textit{data damage} than $S$-invariance a topic which will be explored in future work;
    \item[(iv)]  In practice, we will increase $n_Q$,  and monitor convergence of the $E$ performance figure, as the basis for setting its minimal sufficient value.
\end{itemize}

\subsection{Adult Income data set}
\label{sec:adult}
We now evaluate our method on a benchmark real data set, specifically the {\em Adult\/} income data set \cite{adult}. We assign $s\equiv 1$  to males, and $u\equiv 1$  to subjects with college-level education or above. We restrict the feature space, $\mathbb{X}\subset \mathbb{R}^2$, to age and hours worked per week, since the remaining features are categorical. We omit the continuous features, capital gain and capital loss, since their S-conditional distributions are near-identical before repair \cite{zhou2023group}.

Based on the design principles derived from the simulation study (Section \ref{sec:opcon}), we set $n_R = 10,000$ and $n_A = 35,222$. Note that we choose the ratio, $\frac{n_R}{n_R + n_A}$  higher than for the simulated data (Section~\ref{sec:simulation}) because the Adult data are subject to noise processes and drifts that induce non-Gaussianity and other higher-order behaviour in the underlying marginals, $\mu_{u,s,k}$ (Equation~\ref{eq:emp}). Similarly, we set $n_Q = 250$ to ensure these S-conditional distributions are represented at high resolution while reducing the computational complexity of the repair.   

Table \ref{tab:adult-repair} summarizes clear evidence   that  our distributional OT-repair method can significantly  reduce  $s|u$-dependence  in both the research and archive data. We note the following:
\begin{itemize}
\item[(i)] 
The feature-specific repair metric, $E_{k}$, is smaller for the unrepaired Adult data than for our simulation, as may be seen by comparing the first row of Table~\ref{tab:adult-repair} with that of Table~\ref{tab:sim-repair}. This is because the simulated $(u,s)$-conditional Gaussian components are well separated, whereas the Adult subgroups---segmented by gender and education---are far less so. Note also that the $E_{k}$ reduce by more than 50$\%$ between the research and archive data, suggesting that $\mathbb{X}_R$ provides an incomplete learning resource for $\mathbb{X}_A$, indicative of nonstationarity in these real data. 
\item[(ii)] Nevertheless, our OT-repair method greatly reduces dependence on gender for each educational subgroup,  with a $\sim 4$-fold reduction in $E$ for $\mathbb{X}_R$ and $\sim 3$-fold for $\mathbb{X}_A$. 
\item[(iii)] It is interesting to note that our repair outperforms the geometric repair for the hours/week feature. This is despite the far greater computational load in computing the OT plans, $\pi^*_{u,s,k}$, on $\mathbb{X}_{R,u,k}^3$ in the geometric repair vs.\ $\mathbb{Q}_{u,k}^3$ in our distributional repair. 
\end{itemize}

\begin{table}
    \centering
    \caption{OT-based repairs to quench conditional dependence of the educational groups on gender in the Adult income data set \cite{adult}. Comparison of our distributional OT-based repair method to the geometric OT-based repair in [10]. Lower figures indicate improved repair.}
    \begin{tabular}{l|ll|ll}
        \toprule
        Repair & \multicolumn{2}{c|}{$E_{k}$ (Research)} & \multicolumn{2}{c}{$E_{k}$ (Archive)} \\
         & Age & Hours/Week & Age & Hours/Week \\
         \midrule
         None & 1.108 & 2.700 & 0.546 & 1.311\\          
         Distributional (ours) & 0.339 & \textbf{0.532} & \textbf{0.310} & \textbf{0.367}\\
         Geometric \cite{gordaliza2019obtaining} & \textbf{0.195} & 2.126 & - & - \\
         \bottomrule
    \end{tabular}
    \label{tab:adult-repair}
\end{table}

\section{Discussion}\label{sec:discussion}
The results presented in Section \ref{sec:experiments} demonstrate the promise of our method for off-sample repair of both simulated and real-world data, and show results comparable to the state of the art for on-sample repair. Further work is necessary to evaluate alternative methods for estimation of the $(u,s)$-conditional marginals, $\mu_{u,s,k}$, that are even more data-efficient and can better handle non-Gaussian, nonstationary and/or non-continuous features. A research question also arises in respect of stopping rules for learning of the marginals for the purpose of designing the OT plan. A related issue is the choice of $n_{Q,u,k}$ as a function of the statistics of the  $\mu_{u,s,k}$ marginals, through parametric or non-parametric models for the marginals.
The OT design problem naturally encourages non-parametric learning methods, which can also allow us to address the important generalization to continuous unprotected attributes, $u\in\mathbb{R}^{n_u}$. 

In this paper, we adopted the barycentre (Equation~\ref{eq:baryt}) as the $s|u$-invariant target design, following \cite{gordaliza2019obtaining}, and consistent with the design objective of unregularised OT (Equation~\ref{eq:Wass}). In future work, we will investigate further the trade-off between partial repair \cite{gordaliza2019obtaining} and information loss (damage) in the marginals, using our distributional OT-repair method. However, non-Wasserstein-based target designs should also be considered, particularly when adopting regularised OT {\cite{cuturi2013sinkhorn}.   

Throughout this paper, we have made a series of simplifying assumptions, most notably that the protected attributes ($s|u$-labelling) of the off-sample (archival) data, $\mathbb{X}_A$, are known or can be estimated with low error. Since our OT repair plans, $\pi^*_{u,s,k}$,  are $(u,s)$-indexed, we rely on the accuracy of these labels to ensure that our repairs are optimal and can generalize to  $\mathbb{X}_A$. Provisions in the recent AI Act \cite{AI_act} of the European Parliament allow for the gathering of data with sensitive attributes specifically for the kinds of AI Fairness (AIF) research, certification and de-biasing that this paper addresses. Nevertheless,  it is typical that sensitive attributes, $s\in\mathbb{S}$, are not measured. Therefore, a priority of our future work will be to extend  our distributional OT-repair methods to $s|u$-{\em un}labelled  $\mathbb{X}_A$, as in \cite{chai2022fairness, zhou2023group, elzayn2023estimating}.

Recall that our OT-repairs are designed per feature, i.e.\ we design $|\mathbb{U}|\times |\mathbb{S}|$ OT plans, $\pi^*_{u,s,k}$, for {\em each\/} $ k \in \{1,\ldots, d \}$. Since  $d \gg 1$ in many applications, this presents a significant computational overhead at the on-sample  stage (i.e.\ the OT plan design: Algorithm~\ref{alg:dist_repair}), notwithstanding the savings achievable via the reduced-state interpolated supports, $\mathbb{Q}_{u,k} \times \mathbb{Q}_{u,k}$, as discussed in Section~\ref{sec:internQ}. The dividend in our feature stratification appears at the off-sample stage (Algorithm~\ref{alg:off-sample}), where we may be repairing torrents of archival data (i.e.\  with $n_A$ potentially unbounded). The (static) repair of each coordinate (feature), $x_{A,k}$, in the feature vector, $x_{A}$, avoids the exponential increase (curse of dimensionality) in the size of the interpolation support set, $\mathbb{Q}_u$ (line 5 of algorithm~\ref{alg:off-sample}), as a function of $k$. The associated increase in the   cost of implementing the truncation (i.e.\ quantization) step~5 and randomisation steps~6-9 of Algorithm~\ref{alg:off-sample} are also then avoided. However, this is at the cost of neglecting the intra-feature correlation structure in the $x_{u,s}$. The impact of this on the performance and efficiency of our algorithm is a subtle one, and will be explored in future work. 

Finally, we know from Brenier's theorem \cite{brenier1987decomposition} that Kantorovich OT repair plans (Equation~\ref{eq:kantorovich_emp}) converge to Monge maps as $n_{Q,u,k} \rightarrow \infty$. The impact of this on the performance of OT-based fairness repairs has yet to be fully explored. We suggest that this could improve the {\em individual\/} fairness of the approach, and not only metrics of group fairness such as  conditional independence (Definition~\ref{def:fair}). This is because Monge maps are functions (i.e.\ mass-splitting is obviated),  and so feature-similar points are repaired similarly.


\section{Conclusion}\label{sec:conclusion}
In this paper, we have presented a novel method for off-sample, OT-based data repair, where data not used to design the repair can still be repaired. We demonstrate performance comparable to the state-of-the-art for on-sample repair in a small, simulated setting and on the Adult income data set, and we demonstrate that our method can also be used to repair off-sample points, which is not possible with direct application of current methods.

We present studies of the influence of the operating conditions, $n_R$ (the quantity of research (on-sample) data on which the repair is designed) and $n_Q$ (the resolution of the interpolating supports of the marginal distributions). We demonstrate that the performance of the method converges for $n_R$ as low as 10\% of the overall number of data, $n\equiv n_R + n_A$. We show that good repairs can be designed for $n_Q \ll n_R$, significantly reducing the complexity of the repair operation.

This method has the potential to make fairness repair significantly more accessible to entities which collect data in the real world, since only a small proportion of the data (which we refer to as the research data) need be labelled with their protected attributes, $s\in\mathbb{S}$. The repair plans can then be efficiently applied to unbounded torrents of archival data, assuming that the stationarity assumptions on which the method relies are met.   Efficient fairness-aware data repair of this kind is increasingly important as we move towards the deployment of the AI Act and its requirements for companies to certify the fairness of their models. It also has an important role to play in the repair and exploitation (as training data) of  non-compliant historic (archival) data sets. 


\bibliographystyle{IEEEtran}
\bibliography{IEEEabrv, bibliography}

%
%
%
\end{document}